\documentclass[letterpaper, 10 pt, conference]{ieeeconf}  % Comment this line out if you need a4paper

\IEEEoverridecommandlockouts                              % This command is only needed if 
\usepackage{comment}
\usepackage{caption}
\usepackage{subcaption}
\usepackage{amsmath,amsfonts,amssymb}
\usepackage{multirow}
\usepackage{array}
\usepackage{gensymb}
\usepackage{float}
\usepackage{textcomp}
\usepackage{graphicx}
\usepackage{algorithm}
\usepackage{algpseudocode}
\usepackage{epstopdf}

\usepackage{manfnt}

\usepackage{cite}
\usepackage[colorlinks=true, allcolors=blue]{hyperref}
\usepackage{hyperref}
\usepackage{xfrac}
\usepackage{mathtools}

\title{\LARGE \bf
A Unified Multi-Layer Framework for Skill Acquisition from \\ Imperfect Human Demonstrations}

\author{Zi-Qi Yang  
and Mehrdad R. Kermani% <-this % stops a space
\thanks{Authors are with the Department of Electrical and Computer Engineering, Western University, London, ON N6A 5B9, Canada.
         {\tt\small Email: zyang524@uwo.ca, mkermani@eng.uwo.ca}}%
}

\begin{document}

\maketitle
\thispagestyle{plain}
\pagestyle{plain}

\begin{abstract}
Current Human-Robot Interaction (HRI) systems for skill teaching are fragmented, and existing approaches in the literature do not offer a cohesive framework that is simultaneously efficient, intuitive, and universally safe. This paper presents a novel, layered control framework that addresses this fundamental gap by enabling robust, compliant Learning from Demonstration (LfD) built upon a foundation of universal robot compliance. The proposed approach is structured in three progressive and interconnected stages. First, we introduce a real-time LfD method that learns both the trajectory and variable impedance from a single demonstration, significantly improving efficiency and reproduction fidelity. To ensure high-quality and intuitive {kinesthetic teaching}, we then present a null-space optimization strategy that proactively manages singularities and provides a consistent interaction feel during human demonstration. Finally, to ensure generalized safety, we introduce a foundational null-space compliance method that enables the entire robot body to compliantly adapt to post-learning external interactions without compromising main task performance. This final contribution transforms the system into a versatile HRI platform, moving beyond end-effector (EE)-specific applications. We validate the complete framework through comprehensive comparative experiments on a 7-DOF KUKA LWR robot. The results demonstrate a safer, more intuitive, and more efficient unified system for a wide range of human-robot collaborative tasks.
\end{abstract}

\section{Introduction}
Despite advances in human-robot skill teaching, practical human-robot teaching systems are fragmented, combining isolated sensing, control, and learning components. This leads to solutions that are brittle to context changes and difficult to tune. For widespread use, it is desirable to provide a unified solution that achieves efficiency, intuitiveness, and universal safety, rather than treating them as separate stages.
A key limitation is that most scenarios only learn geometric paths and later add fixed impedance, missing the human's intent for variable force interaction. Furthermore, critical issues like singularity handling are often separated from interaction design, causing unpredictable motions. Safety is also typically limited to the EE, failing to ensure full-body compliance during unexpected contact. Consequently, these systems lack the generalizability needed for real-world tasks.

Current approaches to LfD address the challenge of ensuring quality by filtering and quantifying multiple demonstrations
\cite{coates2008learning, sakr2022quantifying}. The subsequent learning process often employs statistical models like Hidden Markov Models (HMMs) \cite{hovland1996skill}, Gaussian Mixture Models (GMMs) \cite{chernova2007confidence}, or Gaussian Process Regression (GPR) \cite{schneider2010robot} to extract a generalized path. Alternatively, dynamic system-based methods like Dynamic Movement Primitives (DMPs) \cite{chen2017robot} offer a time-dependent framework with convergence guarantees. However, a significant limitation of these existing methods is their inherent reliance on multiple demonstrations and their time-dependent nature. This multi-demonstration requirement increases the user burden, while the time-dependency makes the reproduced paths rigid. This rigidity compromises efficiency and safety, as the EE cannot adapt to unexpected contact or environmental constraints, instead following a predefined temporal path. In contrast, the proposed framework requires only a single demonstration and generates a time-independent reproduction. This key difference enables a more flexible and compliant behaviour, allowing the EE to dynamically adapt its path in response to interaction forces and obstacles. 

\begin{figure*}[!ht]
   \begin{center}
\includegraphics[width=0.88\textwidth]%width=7cm,height=3cm
{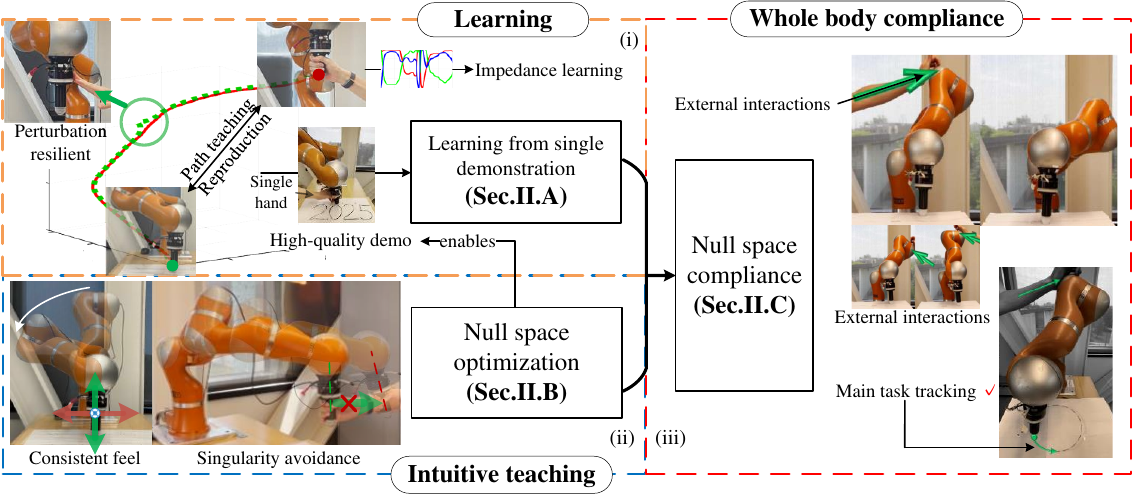}
   \end{center}
   \caption{Framework to achieve intuitive and efficient skill teaching on a KUKA robot in a physical HRI setup: (i) learning from demonstration and path reproduction under perturbation (solid red:  demonstration; dashed green: perturbed reproduction; green dot: start; red dot: end) (ii) null space optimization for pose adjustment (white arrow) and singularity avoidance, enabling high-quality single-handed demonstration of writing “2025”, which is challenging to execute precisely with one arm; and (iii) null space compliance that allows active user interactions (green arrows) without interfering the main task.}
   \label{Control_Diagram}
\end{figure*} 

Secondly, studies indicate that direct user feedback, such as haptic guidance, can offer more intuitive and effective support \cite{liu2006learning}.  A notable approach involves the use of Virtual Fixtures (VF), which have been integrated into demonstration procedures to offer haptic guidance \cite{rosenberg1993virtual}.
Subsequent efforts aimed to incorporate performance constraints to avoid kinematic singularities and joint limits \cite{dimeas2018manipulator
}.
Further, researchers also focused on improving the physical interaction feeling. To  reduce  variability in interaction feel  from varying robot dynamics,  methods were proposed to optimize mass, inertia, and friction parameters, thereby reducing physical effort%petersen2016mass
\cite{petersen2014mass}.  Other works combined VFs with kinesthetic teaching to support tasks with predefined geometry %papageorgiou2021task
\cite{papageorgiou2020passive}, while learning or tracking of manipulability ellipsoids also showed potential for optimizing LfD tasks, though often relying on prior geometric knowledge%rozo2017learning
\cite{jaquier2018geometry}. 
However, these approaches remain limited in addressing the core of key user experience challenges, such as singularity avoidance, dynamic adaptability, and consistent physical interaction. Our method enhances the quality and intuitiveness of the kinesthetic teaching process through a unified null-space optimization framework that proactively manages singularities and ensures a consistent interaction feel across conditions.

Finally, incidental contact or environmental constraints close to the robot body can impact the safety during reproduction. Traditional methods like impedance control \cite{hogan1985impedance} create a spring-like response to perturbations but often result in jerky post-interaction motion due to large positional errors. Subsequent research has improved upon this by adaptively regulating impedance parameters to manage interaction forces \cite{abu2020variable}, or by using human input to adapt robot trajectories \cite{khoramshahi2018human}. Besides, techniques such as optimization \cite{ficuciello2015variable} and hierarchical control \cite{dietrich2019hierarchical} have been employed to manage null-space motion and handle workspace constraints. However, a critical limitation persists: these existing safety strategies are typically localized and cannot be generalized to ensure compliance across the entire robot body. To address this, we introduce a null-space compliance method that enables full-body compliance,  allowing the entire robot to adapt safely to unexpected contact and constraints without degrading main task performance.
{Overall, this work introduces a unified, multilayer framework that combines:}
1) a single-demonstration learning framework with learned impedance for robust, perturbation-resistant motion reproduction; 2) a null-space optimization technique that ensures a consistent interaction feel during teaching and proactively avoids singularities; and 3) a null-space compliance method that provides full-body compliance for safe response to accidental contact, together enabling reliable and intuitive skill acquisition from imperfect demonstrations.
The rest of the work is as follows: Section \ref{II} details the methods, Section \ref{Experimental Evaluation} presents experimental evaluations, and Section \ref{IV} concludes the work.

\section{Methodology}
\label{II}
In this section, we present the three layers that enable robust, compliant LfD built upon a foundation of robot full-body compliance. The overall architecture is illustrated in Fig. \ref{Control_Diagram}. In the following, we present the layers for real-time LfD, null space optimization for intuitive demonstrations, and null space compliance for generalized safety.

In this work, we consider an $n$--DOF redundant serial torque-controlled robot manipulator. The dynamic model of the robot in joint space can be expressed as, 
\begin{equation}
\label{dynamicmodel}
M(q)\ddot q + C(q,\dot q)\dot q + g(q) = \tau_{c} + \tau_{n} + \tau_{f} + \tau_{ext},
\end{equation}
where ${q \in \mathbb{R}^{n} }$ denotes the joint configuration, ${M(q) \in \mathbb{R}^{n\times n} }$ is the inertia matrix,  ${C(q,\dot q) \in \mathbb{R}^{n \times n} }$ represents the Coriolis and centrifugal matrix, ${g(q) \in \mathbb{R}^{n} }$ denotes the gravitational torques, whose effect can be accurately compensated in most
industrial robot’s internal controllers. Moreover, $\tau_{c}$ {and} $\tau_{n}\in \mathbb{R}^{n}$ denote the control torques produced by LfD controller in Sec. \ref{lfd} and null space controller in Sec. \ref{comp}, respectively,  $\tau_{f}\in \mathbb{R}^{n }$ is the joint frictional torques, and
${\tau_{ext}\in \mathbb{R}^{n }}$ represents unknown external interaction torque exerted on the EE reflected on robot joints. The dynamic model \eqref{dynamicmodel} does not include $\tau_{no}\in \mathbb{R}^{n}$ produced by the null space optimization in Sec. \ref{secopt} as it only functions during the demonstration stage.
In the following, the dependency on joint configuration $(q, \dot q)$ is omitted for brevity. 

\subsection{Real-time Learning from Demonstration}
\label{lfd}
We first enable the robot to learn from single-shot demonstrations. This layer consists of three parts, namely a three-dimensional (3D) Fast Diffeomorphic Matching (FDM) to match the demonstrated path, a FDM-based motion generator to estimate reproduction velocity, and an extended Kalman filter (EKF)-based framework to correct the reproduced motion from external perturbations.

The FDM method introduced in \cite{perrin2016fast}, was originally proposed for computing diffeomorphisms in both 2D and 3D settings.  However, its validation was conducted only in $\mathbb{R}^2$, where a diffeomorphism $\Phi$ maps point $x_i\in X \subset \mathbb{R}^2$ to $y_i \in Y \subset \mathbb{R}^2$. In this work, we extend the application of this method to 3D, where $X \subset\mathbb{R}^3$ and $Y \subset\mathbb{R}^3$ are two sequences of distinct points.
Each point in $X$ is transformed using the locally weighted translation,
\begin{align}\label{FDM}
  \hat y_i = \phi_{\rho_j, c_j,v_j}(x_i) = x_i + k_{\rho_j}v_j,
\end{align}
where $\hat y_i$ denotes the translated point in $\hat Y\in \mathbb{R}^3$, and $c_j$ and $v_j \in \mathbb{R}^3$ denote the center and direction of the $j^{th}$ translation, respectively. A smooth diffeomorphism is formed by composing locally weighted translations, where $k_{\rho_j} = e^{-\rho_j^2||x_i-c_j||^2},$
is a Gaussian Radial Basis Function (RBF) kernel. Here, $\rho_j \in [0,\mu\rho_{\max}]$ is an optimizable coefficient that minimizes the distance between $\phi_{\rho_j, c_j,v_j}({X})$ and $Y$,
\begin{align}\label{rho}
   & \rho_j := \min \,\, \frac{1}{N}\sum_j ||\phi_{\rho_j, c_j,v_j}({X}) - Y||^2,
\end{align}
where $N$ denotes the number of points in $X$ and $Y$, and $\rho_{\text{max}}=\frac{e^{ (\sfrac{1}{4})}} {\sqrt{2}||v||}$ as the maximum value for better mapping. In this work, the point set $X$ is a straight line connecting the start and the end of the demonstrated path $Y$. 
First, we select the point $x_m$ in $X$ that is the furthest from the points in $Y$, and update  $c_j:=x_m$  and $\iota:=y_m$  with index $m$. Then, $\rho_j$ is optimized within $[0,\mu\rho_{\max}]$ to minimize  \eqref{rho}. The direction of the $j^{th}$ translation $\phi_{\rho_j, c_j,v_j}$ is updated to $v_j = \beta (\iota - c_j)$, with $0<\mu<1$ representing an adjustable coefficient for $\rho_\text{max}$, $0<\beta<1$ denoting the learning rate, and $K=\,$120 is the total number of locally weighted translations specified by the user, where the loop iterates over each translation. Finally, we update the estimated demonstration path $\hat Y$  by compositing the $K$ locally weighted translations as, $\hat{Y}=\Phi( X^{(K)})=\bigl(\phi_{\rho_1,c_1,v_1}\circ\phi_{\rho_{2},c_{2},v_{2}}\circ\cdots\circ\phi_{\rho_K,c_K,v_K}\bigr)( X)$.

Although the FDM can accurately
match both 2D and 3D paths, the motion generated by the FDM in \cite{perrin2016fast} deviates substantially from the original demonstrations, and its ability to generalize deteriorates in higher-dimensional (3D) and in complex cases. Therefore, we propose to formulate the FDM-based motion generator as, 
\begin{align}\label{DS}
   \dot y_{\text{fdm}}&= -\zeta\,J_{\Phi}\,\hat x,
\end{align}
where $\zeta = \zeta\, + \, \eta\,\Delta \zeta\in \mathbb{R}^{3\times 3}$ is a velocity modulation term. In which, $\Delta{\zeta}= \frac{(\cdot)^T v_{e}}{||(\cdot)^T v_{e}||+\epsilon}$, with $(\cdot)$ denotes the partial derivative $\frac{\partial \dot{{y}}_{\text{fdm}}}{\partial {\zeta}}$, and $v_{e} \in \mathbb{R}^3$ denotes the velocity error between the original demonstration and the FDM-generated velocity,  $J_\Phi \in \mathbb{R}^{3 \times 3}$ denotes the Jacobian matrix of the diffeomorphism $\Phi$, $\hat x \in \mathbb{R}^3$ represents the individual point estimated from $\hat{X}=\Phi^{-1}(Y)$.  $\eta$ and $ \epsilon$ denote the adaptation rate and stability constant, respectively. 
In practice, $\Delta\zeta$ is filtered by a moving average filter and applied with clamping.

The proposed FDM-based motion generation method can improve estimation accuracy, but further improvements are needed to address residual estimation error and post-perturbation correction. We further propose to take advantage of the Kalman gain scheduling feature of the EKF framework to achieve the millimeter-level reproduction accuracy and perturbation recovery. The corrected velocity is,
\begin{align}\label{v_ekf}
    \dot y_{\text{ekf}} = \dot y_{\text{fdm}} + b.
\end{align}
It is composed of the FDM velocity with an additive velocity bias $b \in \mathbb{R}^3$ that the EKF estimates online. Different from the common EKF, which calculates the estimation residual from the actual measurement, we update the estimation residual using the original demonstration.

Additionally, the robot impedance is learned from the demonstration to parameterize its interaction feel and response to perturbations across different task phases.
Inspired by \cite{kronander2015passive} and building on the
approach introduced in \cite{yang2025nullspace}, we formulate a velocity tracking controller, $F_c = D(*) \, \dot p_c,$
where $\dot p_c =  \dot y_{\text{msr}}-\dot y_{\text{ekf}}$, and 
$\dot y_{\text{msr}}$ denotes the actual translational velocity measured from the robot. The control force $F_c \in \mathbb{R}^{3}$ is further converted to joint control torque $\tau_c \in \mathbb{R}^{n}$ through the robot Jacobian matrix $J \in \mathbb{R}^{6 \times n}$, where $F_c$ is first augmented with zeros for the orientation components. A separate PD controller maintains the EE
orientation, keeping it pointed toward the ground. $ D(*) \in \mathbb{R}^{3\times 3} $ is the velocity direction parameterized damping, it is designed to selectively tune the dissipation of energy in desired and undesired directions. 

In this work, the goal is to react to arbitrary external interactions with compliance while maintaining the tracking of demonstrated paths. Thus, let $*:=\dot y_{\text{ekf}}$, the parameterized damping is defined as, $D(*)=U(*)\,\Xi \, U(*)^T,$
where matrix $U(*) =[\hat e_1, \hat e_2, \hat e_3]  \in \mathbb{R}^{3\times 3}$ contains the estimated orthonormal principal axes $\hat e_1, \hat e_2, \hat e_3$, with $\hat e_1 \in \mathbb{R}^{3}$ pointing in the desired direction of motion, in that $\hat e_1 := \sfrac{*}{||*||}$, and the rest are represented as $\hat{ e}_{t} :=  \frac{\mathbf v^{(t)}}{\max(\|\mathbf v^{(t)}\|,\varepsilon)}$ with  $t=2,3$. Moreover, $\mathbf v^{(t)} :=  e_k -\bigl(\hat{ e}_t^{\,T} e_k\bigr)\hat{ e}_t,$ where  $e_k$ denotes the $k^{th}$ canonical axis. For the $1^{st}$ canonical axis, $e_1 = [1, 0, 0]$, $\varepsilon \in \mathbb{R^+}$ denotes a numerical stability constant.  Additionally, sign continuity of the components in $U$ is ensured in practice. $\Xi = \text{diag}(\xi_1,\xi_2,\xi_3)\in \mathbb{R}^{3\times 3}$ is a user-defined diagonal matrix with non-negative elements (typically 10--100), tuned according to the perceived interaction feel. This is to adjust the perceived interaction behaviour. For example, setting $\xi_1 = 0$ and $\xi_2, \, \xi_3>0$ allows for resisting the external forces that are not aligned with its direction of movement. In practice, a surface contact force (e.g., for polishing) can  be generated by adding an extra term to \eqref{v_ekf}, $\dot y_{\text{ekf}} = \dot y_{\text{mod}} + \dot y_{\text{c}}+ b,$
where $\dot y_{\text{c}} = \alpha_c \hat{ e}_3$ creates an adjustable surface-directed velocity.

\subsection{Intuitive Teaching with Null Space
Optimization}
\label{secopt}
In the previous section, we introduced the layer enabling single-shot LfD. Building on the approach in \cite{yang2026user} and to further enhance teaching quality and intuitiveness, we propose a null space optimization layer for the teaching stage.
This method is separated into a null space optimization to address inconsistent interaction feeling and manage singularities, and a variable Cartesian impedance controller for the compensation of residual variations in the perceived interaction forces.

The optimization method includes three metrics: the directional manipulability, the apparent inertia unification, and the elbow singularity avoidance. The first metric works during the initial stage and has less dominant weight. It mildly suppresses the excessive anisotropy of the manipulability ellipsoid, reflecting how easily the user can initiate the movement along different directions. It is formulated as, $\mathcal{C}_1  = \sqrt{u^T \;\Upsilon \;u},$
where $\Upsilon =J_{v}  J_{v}^T \in\mathbb{R}^{3\times 3}$ represents the manipulability matrix for translational motion, with $J_v \in \mathbb{R}^{3\times n}$ denotes the translational part of the full Jacobian matrix.  $u$ denotes the unit vector coincident with one of the base frame axes. For example, $u = [0,0,1]^T$ when the corresponding dominant axis of the manipulability ellipsoid's largest projection on the base frame is along the $z$--axis.  The EE orientation is also set to a constant level as in the previous subsection.

Since the inertial properties of the robot are an essential factor in maintaining the force consistency during interactions. The second metric works as a dominant term to consider the real-time inertia distribution, regulates the apparent inertia along selected directions to be approximately the same using the proposed modified dynamic conditioning index (mDCI). It is formulated as, $\mathcal{{C}}_2 \;=\; \frac{1}{2}\,{\Bar{E}}^T({q})\,{\Bar{W}}\,{\Bar{E}}({q}),$
where  $\Bar{W}= \mathrm{diag}\{\Bar{\mu} I_3\}$ is a diagonal weighting matrix with $\Bar{\mu}\geq 1$, $\Bar{E}(q)  = [\Delta_1 - \Bar{\delta},  \:\Delta_2 - \Bar{\delta}, \:\Delta_3]^T$, in that $[{\Delta_1},\, \Delta_2,\, \Delta_3]\;=\;\mathrm{sortDescending}\bigl(\lambda_{11},\, \lambda_{22},\, \lambda_{33}\bigr)$. $\Bar{\delta} = \frac{1}{2}\,\bigl(\Delta_1+\Delta_2\bigr)$, in that $\{\lambda_{11},\lambda_{22},\lambda_{33}\}$  denote the diagonal elements of current apparent inertia matrix $\Lambda$.
In other words, the selected directions are the two principal axes with dominant apparent inertia to avoid incurring frequent internal motion caused by impractically optimizing apparent inertia along all principal axes towards their average. To further enable true consistent interaction force along every principal direction, a variable Cartesian impedance controller can be introduced by taking advantage of the perceptual damping ratio $\frac{D_\text{var}}{\Lambda}=\frac{D_d}{\Lambda_d}$, where $D_{\text{var}}\in \mathbb{R}^{3\times 3}$ is the variable damping matrix, and $D_d, \Lambda_d\in \mathbb{R}^{3\times 3}$ denote the desired damping (set to render a comfortable
interaction force range of 6–7 N \cite{cirillo2015conformable}) and desired inertia (chosen based on the load), respectively.

The optimization using the above two metrics  renders a consistent interaction feeling. To further reduce users' cognitive load, avoiding getting stuck in a singular configuration when ignoring robot limitations, we focus on the most common elbow singularities due to the overextension of the robot. It is the last part of the optimization layer, formulated using the Local Conditioning Index (LCI) as, $\mathcal{C}_3 =  \frac{r_{\mathrm{min}}\!\bigl(\Upsilon\bigr)} {r_{\mathrm{max}}\!\bigl(\Upsilon\bigr){+\epsilon}},$
where $r_\mathrm{max},\, r_\mathrm{min}$ denote the largest and the smallest eigenvalue of the manipulability matrix $\Upsilon$. Unlike the conventional LCI metric that relies on $J_v$, we employ the manipulability matrix so that eigen decomposition can be used in place of a full singular value decomposition (SVD), thereby reducing computational cost. In addition, to prevent the gradient $\nabla_{q}\mathcal{C}_3$ from becoming discontinuous when eigenvalues coincide, a small positive bias $0<\epsilon\ll r_\mathrm{max}$ is introduced in the denominator.  Further, to avoid continuously maximizing the $\mathcal{C}_3$, we apply a switching strategy to its gradient: we scale $\nabla_q \mathcal{C}_3$ by a factor $a$ when $r_{\min} < r_{\mathrm{thr}}$, and set it to zero otherwise.
Here, $r_{\mathrm{thr}}$  is a user-defined, hardware-dependent threshold that that controls when the switch occurs.
$a = 
  \mathbf{K}_p\,\Delta_{r}
  \;+\;
  \mathbf{K}_d\,\dot{\Delta}_{r}$, in that
$\Delta_r = r_{\mathrm{thr}}-r_{\min}\in\mathbb{R}$, $\mathbf{K}_p\,, \mathbf{K}_d\in\mathbb{R}^+$ are proportional and derivative gains of a PD controller that adaptively modulate how strong the repulsion is when approaching singularities. In practice, $K_p$ determines the repulsion strength relative to proximity to the threshold, while $K_d$ damps rapid variations to prevent oscillations.

Finally, the optimization can be formulated as, $\mathcal{C} = -w_1\,\mathcal{C}_1 + w_2\,\mathcal{C}_2 - w_3\,\mathcal{C}_3,$ then write,
\begin{align}\label{cost}
   &\min_{q} \quad  \mathcal{C}(q),
\end{align}
which is solved analytically. $w_1, \,w_2$ and $w_3$ are tunable weights (bounded between 1--5) for the directional manipulability optimization, apparent inertia regulation, and singularity avoidance features. To avoid affecting the main task, we project $\nabla_q \mathcal{C}$ using the dynamically consistent null-space projector \cite{khatib2003unified}, $N_{\text{dyn}} \;= I-J_v^T \Bar{J}_v^T,$
where $I \in \mathbb{R}^{n\times n}$ denotes an identity matrix.
$\Bar{J}_v =  {M}^{-1}\,J_v^T \, \Lambda \in \mathbb{R}^{n\times 3}$ is the dynamically consistent generalized inverse of $J_v$.
Consequently, the null space control torques $\tau_{no}\in \mathbb{R}^{n}$, combining the gradient of \eqref{cost} with  feedforward joint vicious-friction  compensation  and  damping, are given by, $ \tau_{no} =  N_{dyn} \Bigl(\alpha_{ns} \nabla_{{q}} \mathcal{C}+\alpha_f \dot{q} - k_D\dot{q}\Bigr),$
where {$\alpha_{ns}$, $\alpha_f\in \mathbb{R}^+$ are empirically tuned null space and frictional gains (typically 1 and 0.1), and $k_D \in \mathbb{R}^+$ is a user-selected damping gain within 1--3 $\frac{\text{Nm}\cdot \text{s}}{\text{rad}}$ to damp the joint motion during optimization for a mild user experience.}

\subsection{Generalize Safety with Null Space Compliance}
\label{comp}
We have addressed learning, reproduction, and user experience above. During  reproduction, EE can respond compliantly to and recover from  external perturbations. 
To generalize safety to the entire robot body to compliantly adapt to external
interactions without compromising the main task performance, we introduce the null space compliance layer. By exploiting redundancy,
the robot body can respond to external interactions compliantly
while preserving
Cartesian motion, effectively dissipating external energy
exerted on the robot body in the null space.
\begin{figure*} [t]
   \begin{center}
\includegraphics[width=0.88\textwidth]
{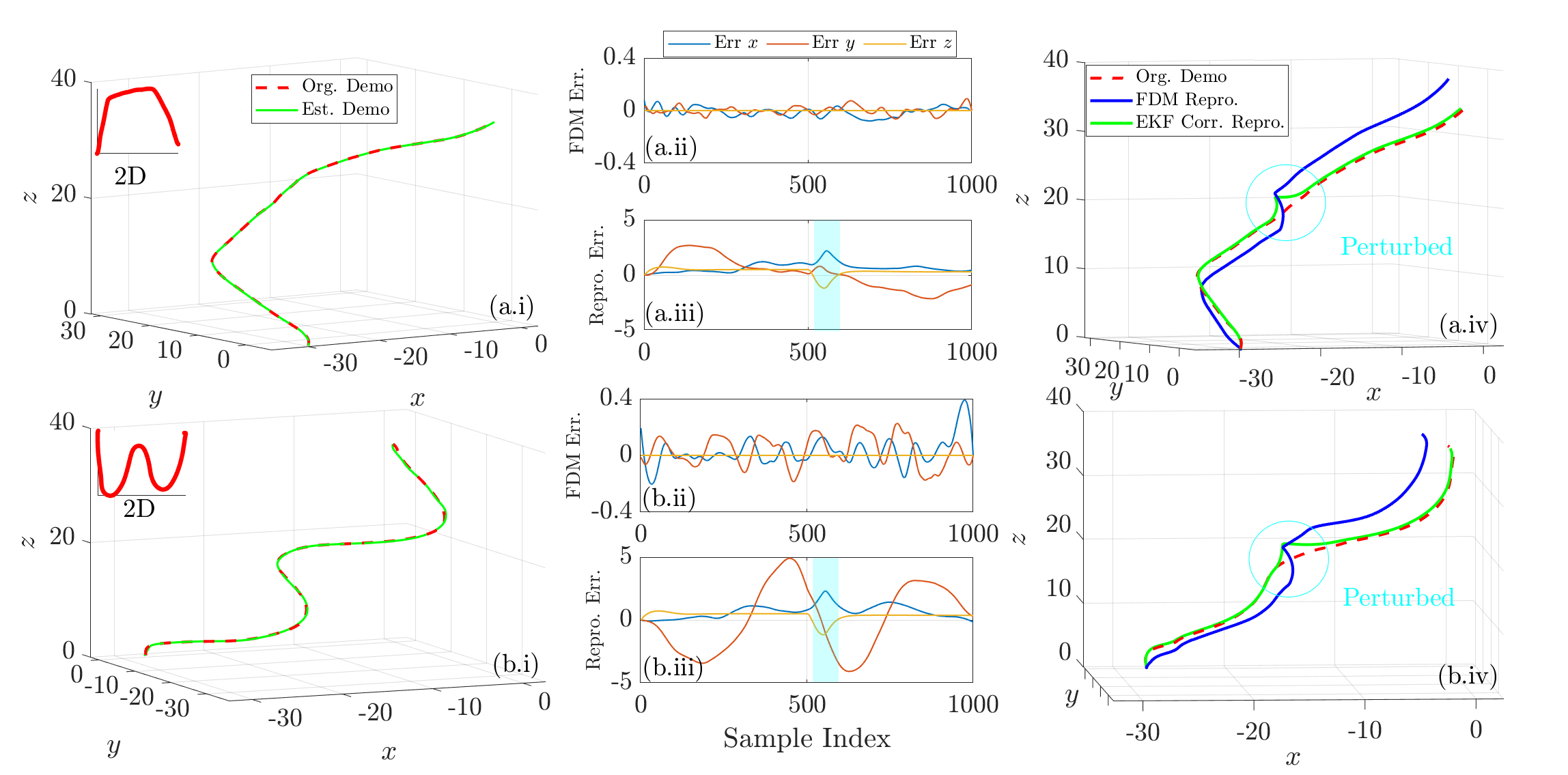}
   \end{center}
   \caption{Experiment A: (a,i) 3D FDM learned path (with a 2D inset for easier visualization), (a.ii--a.iv) estimation error of the learned path, reproduction error, and reproduction under perturbation for a trapezoid-shaped path; (b.i–b.iv) corresponding results for a `W'-shaped path. Light blue shaded areas indicate the period of interaction, with the corresponding impact highlighted in circles. All values are expressed in centimeters, and all paths start at zero height ($z$-axis).}
\label{FDM}
\end{figure*} 
\begin{figure} [t]
   \begin{center}
\includegraphics[width=8.0cm]{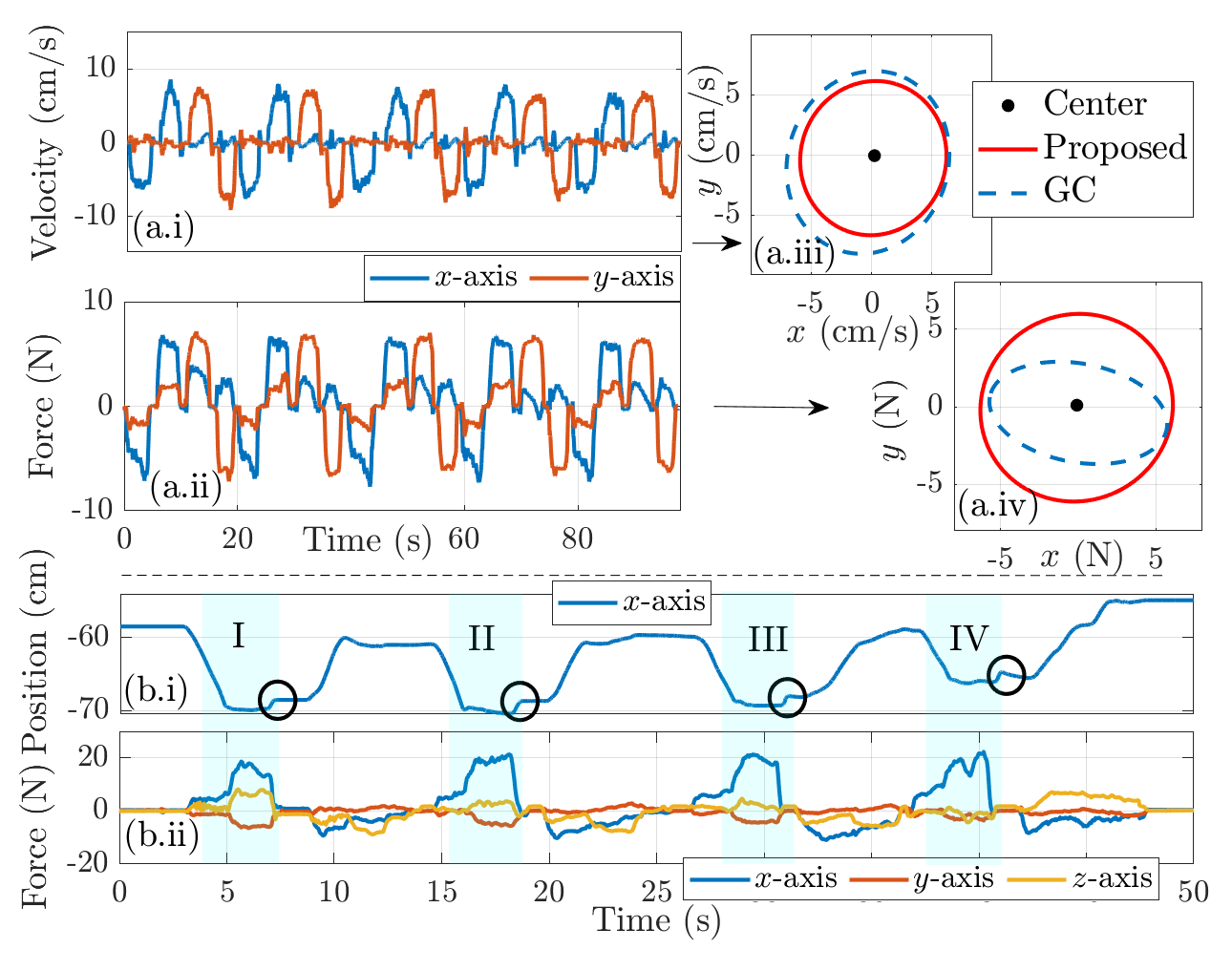}
   \end{center}
   \caption{Experiment B: (a) Measured velocity and force for an L-shaped path using the proposed method (i-ii) and consistency comparison with the GC method, expressed in isotropic representation (iii-iv). (b) Singularity avoidance test showing $x$-axis position (i) and interaction force (ii); shaded areas indicate interaction, with release causing backlash (circled).}
\label{opt}
\end{figure} 
\begin{figure} [t]
   \begin{center}
\includegraphics[width=8.0cm]{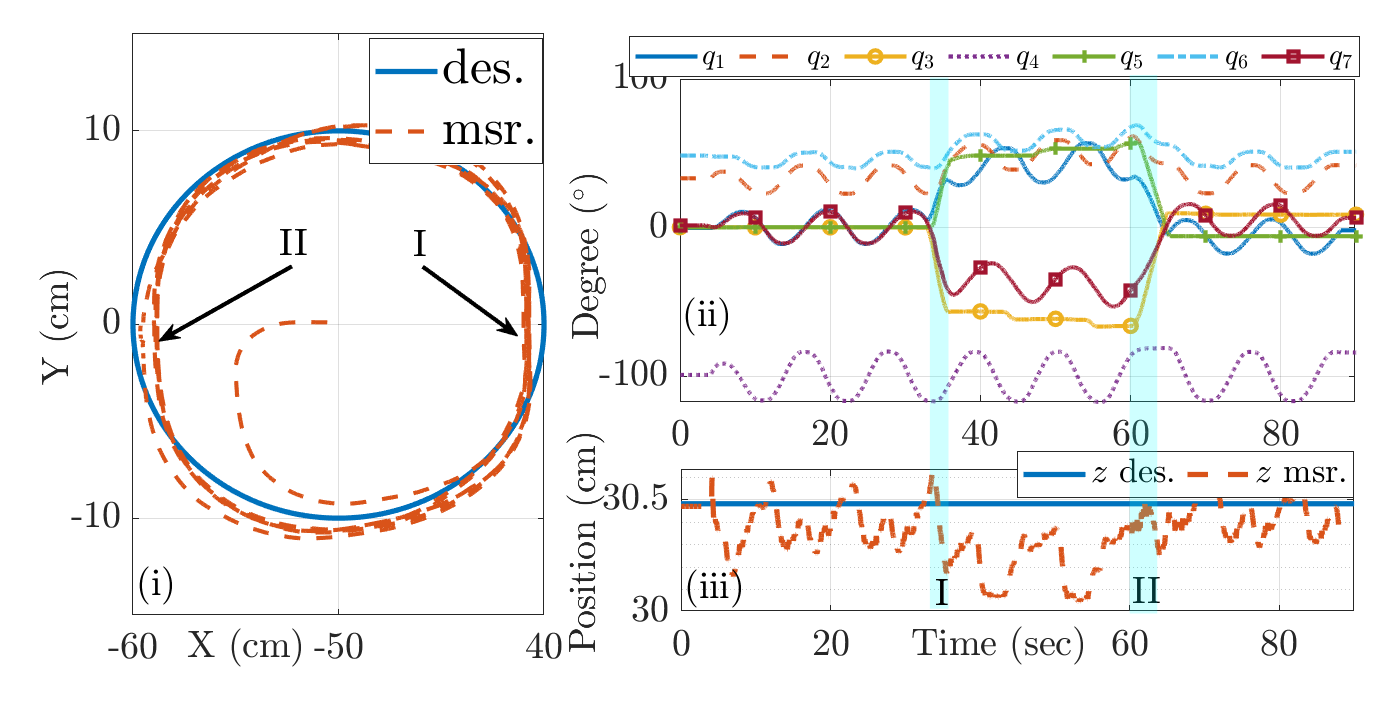}
   \end{center}
   \caption{Experiment C: (i) tracking desired path in the XY plane, with external interactions' impacts labelled with I and II; (ii) joint deviations due to external interactions. (iii) position variation along the $z$-axis. Light blue shaded areas indicate the period of interaction.}
\label{compliance}
\end{figure}  

Let the relation between measured EE velocity $\dot x_{\text{ee}} \in \mathbb R^6$ and measured joint velocity $\dot q \in \mathbb R^n$ be expressed as $\dot x_{\text{ee}} = J(q) \dot q$. In the following, the dependency on $q$ is omitted for brevity.
A redundancy solution ($n > 6$) to it is achieved through a null space projector $\dot q = J^\dagger \dot x_{\text{ee}} + N \dot q$,
where $J^\dagger = J^T(JJ^T)^{-1} \in \mathbb R^{n\times 6}$ denotes the pseudoinverse of $J$. $N \in \mathbb R^{n\times n}$ is a null space projection matrix of full Jacobian matrix $J$, denoted as $N=I-J^\dagger J$,
where $I \in\mathbb R^{n\times n}$ is an identity matrix.  By utilizing the null space projection, redundant DOFs can be exploited to move within the null space of $J$.
This implies that the projection can be potentially used to project the external interaction forces to the null space. 
This is particularly valuable in scenarios where the robot interacts with its environment.
In light of this, we propose a novel null space impedance control law as,
\begin{equation}
\label{taun}
  \tau_n = N d_n \alpha_f \dot q_{d} + N{\alpha_d}(\dot q_{d} - \dot q),
\end{equation}% -\tau^{ext}
where $d_n \in \mathbb R^+$ represents the null space damping parameter. $\dot q_{d} = J^\dagger \dot{{y}}_{\text{ekf}}
\in \mathbb R^{n}$ denotes the desired joint velocity converted from the reproduction velocity derived in \eqref{v_ekf}, and $\dot q$ represents the actual joint velocity. To maintain the correct matrix structure, the orientation component is augmented to $\dot{{y}}_{\text{ekf}}$ with a value of zero. In the experiments, a separate PD controller maintains the EE orientation, keeping it pointed toward the ground. The first term, $Nd_n \alpha_f \dot q_{d}$, functions as a feedforward compensation term for joint frictional torques, where $\alpha_f$ denotes the frictional gain parameter. By using the desired joint velocity \(\dot{q}_d\), which provides a cleaner signal, and the null space projector \(N\), the compensation is applied smoothly to the redundant DOFs. This enhances the joint responsiveness to external interactions.
The second term $N \alpha_d(\dot q_{d} - \dot q)$ utilizes null space projection to extend the joint motion into the null space of Cartesian space main tasks (e.g. polishing, buffing). Here, \( \alpha_d \) represents the damping gain parameter that regulates the level of compliance. This term enables the redundant DOFs to respond compliantly to unknown external torque $\tau_{ext}$. The deviation between \( \dot{q} \) and \( \dot{q}_d \) caused by the external force generates damping torques through this term. The damping effect effectively dissipates external energy, reducing the impact on main task tracking performance.  Combining the two terms, we ensure external energy is dissipated through the joint damping effect, without being expended in resisting joint friction.  Moreover, the system’s energy dissipation under the influence of external interaction forces has been analyzed in \cite{yang2025nullspace}.

\section{Experimental Evaluation}
\label{Experimental Evaluation}
To evaluate the proposed three layers, we conducted experiments on a 7-DOF KUKA LWR IV+ robot manipulator as shown in Fig. \ref{Control_Diagram}. 
The control algorithms are implemented on a remote Ubuntu PC with the Robot Operating System (ROS) framework. This remote PC establishes communication with the KUKA robot controller via UDP, utilizing the Fast Research Interface (FRI) with a sampling rate of 200 Hz. A $6$--axis ATI Gamma F/T sensor is attached to the EE for reference purposes. Throughout the experiments,  the desired EE orientation is configured as $[\pi,0,\pi]^T$  (rad) to point towards the ground, and is regulated by a PD controller with gains tuned using the Ziegler–Nichols method. The robot is controlled in joint impedance mode, in which the controlled torque is the only input to the robot, with
the control mode’s stiffness and damping interfaces disabled. Proper compensation for gravitational torques is ensured, including the effects of the F/T sensor and a lightweight 3D-printed tool. {A handheld ATI F/T sensor is used in robot body interaction experiments for reference.}

\subsection{Experiment A: Learning, Reproduction and Perturbation}
In this experiment, we evaluate the learning and reproduction accuracy using reference paths from the LASA Handwriting dataset \cite{KhansariZadeh2010LASA}, with the original 2D dataset extended into the full 3D Cartesian space. External perturbations are introduced to evaluate anti-perturbation and compliance capabilities. The learning and reproduction results for trapezoid and `W'-shaped paths are shown in Fig. \ref{FDM}.a and b, respectively. Subfigures i and ii show that the FDM estimation errors remain negligible, staying below 0.3 cm. In subfigures iii and iv, the corrected reproduced paths (green)  closely follow the demonstrations (red), with errors generally under 2 cm.  Perturbation recovery and compliance are validated in subfigure iv, where the affected regions are highlighted with circles. In comparison, the FDM velocity generated by \eqref{DS} (dark blue) yields unsatisfactory reproduction accuracy and fails to recover under perturbations. The experimental process is shown in Fig. \ref{Control_Diagram}.i.

\subsection{Experiment B: Null Space Optimization}
In this experiment, we evaluate the null space optimization layer. First, the variations in null space configuration during the initial optimization phase are illustrated in Fig. \ref{Control_Diagram}.ii (indicated by the white arrow). After the initial phase, we examine force consistency along each direction. The user was instructed to trace an ‘L’–shape path five times in both the XY and YZ planes. Each segment was approximately 20 cm in length and aligned with one of the principal axes of the base frame, while the overall position of the path was chosen randomly.
The ‘L’–shaped path ensures that data can be clearly associated with different Cartesian directions, enabling a more detailed analysis of interaction force consistency.  Throughout these movements, the velocity was maintained consistently to the best of the tester’s effort. Fig. \ref{opt}.a.i--ii show the raw measurements of the interaction velocity and force, which are further converted into isotropic representations in Fig. \ref{opt}.a.iii--iv for clearer interpretation. As illustrated, the proposed method (red) delivers a more isotropic oval shape, indicating improved consistency of interaction forces across directions compared to the gravity compensation (GC) method in \cite{papageorgiou2021task}. Results for the YZ plane are omitted for brevity. Elbow singularity is examined by pulling the EE away from the base approximately along the $x$-axis, as shown in Fig. \ref{Control_Diagram}.ii (right). A resistant
force of around 20 N was generated to counterbalance the user’s pulling force, as shown in Fig. \ref{opt}.b.ii, thus blocking further extension, as shown in Fig. \ref{opt}.b.i.

\subsection{Experiment C: Null Space Compliance}
In this experiment, we investigate the efficacy of the null space compliance control layer introduced in subsection \ref{comp}. We aim to demonstrate that the proposed null space controller enables the redundant DOFs to respond compliantly to unknown external interactions,
thereby generalizing safety to the entire robot body.  The robot tracks a circular reference path with a radius of 10 cm, centered at (-50, 0, 30.5) cm. The choice of a circular path is to highlight the effect of external forces, as its constant curvature and continuous directional change make any tracking deviations due to disturbances easily observable and measurable. External interaction forces are inherently sudden and unpredictable. For reproducibility, two abrupt and randomly occurring impacts of similar magnitudes  ($\approx$ 27 N), measured by a handheld F/T sensor as reference, were applied in the opposite direction to the 4$^{th}$ joint (green arrows), as shown in Fig. \ref{Control_Diagram}.iii.  Their impacts on the main task tracking and joint angles are illustrated in Fig. \ref{compliance}.i and ii. labelled with I and II. As shown in (i), the main task was minimally affected. This is attributed to the joints having compliantly deviated ($\approx $50$^\circ$) from their initial positions, efficiently dissipating external energy within the null space, as shown in (ii). The robot’s responsiveness is also enhanced by the compensation for joint friction. Finally,  subfigure (iii) indicates that the contact with the surface is also not impacted by the interactions, with minor fluctuations observed under 5 mm.

\section{Discussion and Conclusion}
\label{IV}
We acknowledge several limitations. First, we do not provide a systematic ablation study or a comprehensive sensitivity analysis of the key hyperparameters. While our current settings follow common practice and were tuned empirically, a more thorough evaluation of hyperparameter effects is left for future work. Second, our experimental validation is conducted on relatively simple path scenarios, and comparisons are limited. Extending the evaluation to more complex environments and benchmarking against a broader set of state-of-the-art methods will be important directions for future study, along with integrating our previous work on magneto-rheological actuation \cite{yang2023computationally, kermani2023antagonistic}.

This three-layer framework facilitates an intuitive and efficient methodology for robotic skill acquisition. It constitutes a versatile HRI platform applicable beyond EE-specific tasks. The proposed method requires neither external force measurement nor estimation and relies only on a limited knowledge of the robot's dynamics. Extensive experimental validation on a KUKA LWR IV+ robot confirmed the framework's effectiveness and practical feasibility.

\bibliographystyle{IEEEtran}
\bibliography{bib}
\end{document}